\documentclass[11pt]{article}

\usepackage[]{acl}

\usepackage{times}
\usepackage{latexsym}

\usepackage[T1]{fontenc}
\usepackage[utf8]{inputenc}

\usepackage{microtype}

\usepackage{inconsolata}

\usepackage{graphicx}

\usepackage{xcolor}
\usepackage{color}
\usepackage{soul}
\usepackage{amsmath} 
\usepackage{url}
\usepackage{amssymb} 
\usepackage{booktabs}
\usepackage{multirow}
\usepackage{graphicx}
\usepackage{tabularx}
\usepackage{caption}
\usepackage{tabularx}
\usepackage[table]{xcolor} 
\definecolor{perf-pink}{HTML}{FFD9D9}
\definecolor{cost-blue}{HTML}{D9E9FF}
\title{Beyond Query Memorization: Large Language Model Routing with Query Decomposition and Historical Matching}

\author{\textbf{Bo Lv}\(^{1,2}\), \textbf{Jingbo Sun}\(^{2}\) 
\\
\(^1\)Tencent Hunyuan,  
\(^2\)University of Chinese Academy of Sciences \\
\\
\texttt{lvbo19@mails.ucas.ac.cn} \\
}

\begin{document}
\maketitle
\begin{abstract}
Optimizing the trade-off among predictive performance and computational cost is a central focus in the deployment of Large Language Models (LLMs). Current routing methods primarily rely on direct mapping from queries to models based on surface-level features, making them susceptible to the memorization trap and leading to poor generalizability on out-of-distribution (OOD) data.
In this paper, we propose DecoR, a novel routing framework that recasts the routing task as a matching process of sifting similar queries from historical logs, effectively mitigating the memorization trap.
To enhance matching accuracy, we introduce a query capability deconstruction method that decouples linguistic surface forms from task-intrinsic requirements, directing matching toward capability dimensions to ground decisions in essential task attributes. Furthermore, we develop CodaSet, a comprehensive benchmark for assessing routing generalization, where experimental results demonstrate that DecoR maintains superior accuracy while substantially lowering inference costs across both in-distribution and OOD settings. All the codes and data are available at \url{https://github.com/lvbotenbest/DecoR}.

% In this paper, we propose DecoR, a routing framework based on query decomposition and historical matching. DecoR leverages a dedicated decomposition model to explicitly deconstruct raw queries into structured capability requirements and domain knowledge. By matching these fine-grained profiles within a historical experience pool, a reinforcement learning-based discriminator evaluates the candidates to determine the optimal model selection. To ensure system reliability and minimize latency overhead, DecoR incorporates an active fallback mechanism that automatically invokes high-performance models when matching confidence is insufficient.

% Experimental results demonstrate that DecoR significantly enhances routing generalizability and effectively mitigates overfitting while maintaining superior cost-effectiveness and inference efficiency.

\end{abstract}

\section{Introduction}
In the practical deployment of Large Language Models~\cite{qwen3_235,deepseekv31}, user queries exhibit significant variance in complexity, implying that not all tasks necessitate the involvement of massive-scale models. 
To this end, Model Routing~\cite{shnitzer2023large,routerbench} has gained prominence as a key paradigm, which dynamically selects appropriate model sizes based on query characteristics to optimize the trade-off between predictive performance and computational cost.

% ----------- pic:verify_in_line  -----------
\begin{figure}[!t]
\centering
\includegraphics[width=0.99\linewidth]{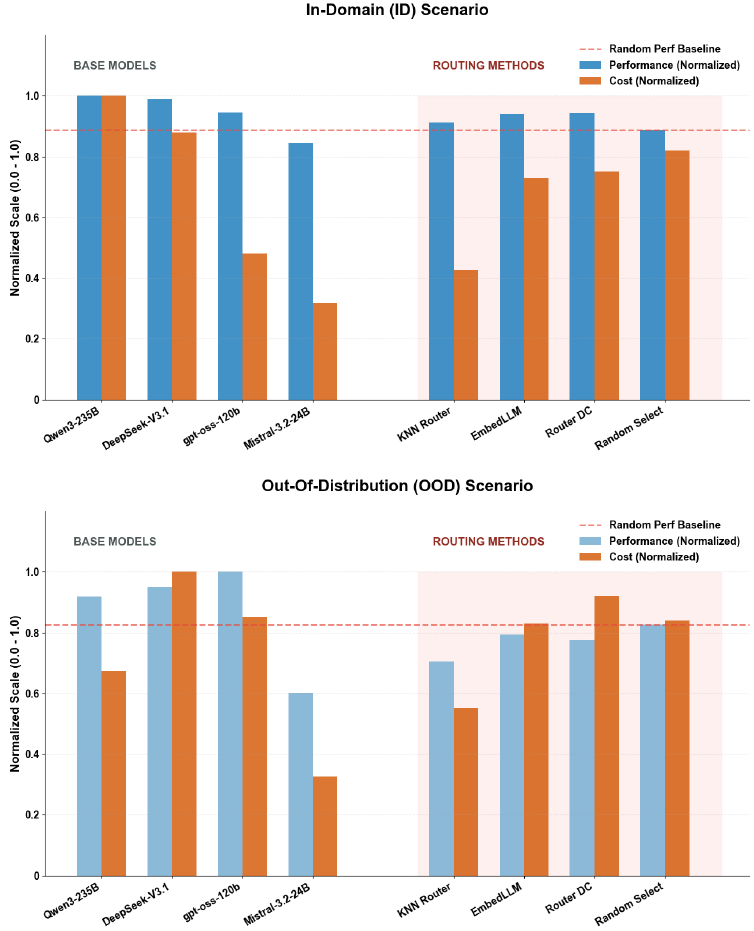}
\caption{Comparison of routing performance and cost in ID  and OOD  settings.}
\vspace{-4pt}
\label{pic:id_ood}
\end{figure}
% ----------- pic:verify_in_line  -----------

However, existing routing methodologies~\cite{RouterDC,HybridLLM,embedllm} predominantly simplify the process into a black-box matching task, establishing direct mappings from inputs to model IDs.
% Such mechanisms are highly susceptible to the Memorization Trap. Specifically, routers~\cite{ZOOTER,routellm} tend to over-rely on surface-level semantic features rather than excavating the deep-seated capability requirements of the task. 
Such mechanisms are prone to a memorization trap, where routers over-rely on surface-level semantics rather than discerning the underlying capability requirements.
Consequently, while these methods excel in training-aligned in-domain (ID) scenarios, their generalization collapses on out-of-distribution (OOD) data. As illustrated in Figure \ref{pic:id_ood}, when tested on OOD tasks, existing routing methods~\cite{RouterDC,routerbench}  fail to surpass the random selection baseline while incurring disproportionately high costs.
Furthermore, since these methods~\cite{routellm,RouterDC,embedllm} are typically trained end-to-end for specific model pairs, their decision logic is tightly coupled with the existing model pool. Any update to the underlying models necessitates costly retraining, incurring additional computational and temporal overhead.

To address these challenges, we propose DecoR (\textbf{Deco}mposition-based \textbf{R}outing), a novel routing framework. 
Departing from black-box mapping, DecoR decomposes queries into capability requirements to match relevant historical query-response logs, leveraging the model’s performance in those matched instances to estimate its prior probability of success for the current query.
Specifically, the framework first employs a \textbf{Query Deconstruction Stage} to decompose the user query into a structured Capability Profile, encompassing Skills ($S$), Knowledge ($K$), and Difficulty ($D$).
Subsequently, the system utilizes the Capability Profile as a core index to identify representative historical query-response logs through a \textbf{Hierarchical Sifting Stage} , ensuring these logs are highly aligned with the user query’s capability requirements. The underlying rationale is that a model’s successful resolution of these matched instances theoretically demonstrates its capability to handle the current query.
Finally, the \textbf{Empirical Decision Stage} identifies the optimal model by balancing performance and cost within the filtered logs. If sifting yields no matched logs, a fallback safety net deploys a high-performance default model to prevent decision failures in OOD scenarios where prior knowledge is unavailable.

To accurately evaluate routing performance in complex semantic environments and ensure fair comparison, we introduce CodaSet (\textbf{C}apability-\textbf{O}riented \textbf{D}ataset for \textbf{A}daptive Routing), a new benchmark constructed using frontier models. Our contributions are as follows:

\begin{itemize}
    \item We propose DecoR, an innovative routing framework that recasts the routing task into a matching process of sifting similar queries from historical logs, thereby effectively mitigating the memorization trap and enabling system iteration solely through log updates without model retraining.
    
    \item We introduce a query capability deconstruction method that decouples linguistic surface forms from task-intrinsic requirements, directing log sifting toward capability dimensions to ground decisions in task attributes and enhance matching accuracy.
    
    % By integrating capability decomposition with an Active Fallback mechanism, DecoR precisely assesses matching confidence in complex and OOD scenarios, significantly improving routing accuracy while ensuring system robustness.
    
    \item We develop CodaSet, a comprehensive benchmark for assessing the generalization of routing systems. Experimental results show that DecoR consistently maintains superior accuracy while substantially lowering inference costs across both ID and OOD settings.
    
\end{itemize}

\section{Related Work}
The rapid rise of various Large Language Models (LLMs) has spurred the development of model routers~\cite{shnitzer2023large,routerbench}, which direct simple queries to smaller models to reduce computational costs without compromising overall performance.
\citet{ZOOTER} proposed a reward-guided routing method that distills reward signals into a routing function to dispatch queries to models with the corresponding expertise.
HybridLLM~\cite{HybridLLM} utilizes a trained language model as a router to dynamically assign queries to either a small or a large LLM based on predicted task difficulty.
Despite its effectiveness, this approach is limited to a binary selection between two models.

To address this constraint, \citet{routerbench} proposed kNN-Router, a framework that estimates model performance by averaging the outcomes of the $k$ nearest training examples, thereby routing each query to the most suitable LLM from a broader pool.
Other recent works have shifted toward representation learning; for instance, \citet{RouterDC} introduced a dual contrastive learning-based router that jointly optimizes query and model embeddings by aligning queries with compatible models and clustering them semantically. 
Similarly, \citet{embedllm} developed an encoder-decoder framework to learn compact embeddings for predicting model-query compatibility via a binary cross-entropy objective. Alternatively, \citet{iclrouter} utilized in-context vectors to capture model capabilities, leveraging the relationship between these vectors and query embeddings to predict a model's performance on new queries.
In addition, \citet{SAT} explored a text-based approach, transforming candidate model performance into textual descriptions and leveraging a trainable LLM to process these features for dynamic selection.

Despite the success of these methods, most existing approaches rely heavily on learning fixed mappings between query embeddings and model representations. This paradigm risks falling into a memorization trap, where the router tends to memorize specific training queries rather than generalizing the underlying relationship between task characteristics and model capabilities.
To overcome this, we propose DecoR, a framework that recasts the routing task as a matching process of sifting similar queries from historical logs, effectively mitigating the memorization trap. This matching is driven by a capability deconstruction method, which decouples linguistic surface forms from task-intrinsic requirements to significantly enhance accuracy.
\section{Method}

\begin{figure*}[!t]
\centering
\includegraphics[width=0.99\linewidth]{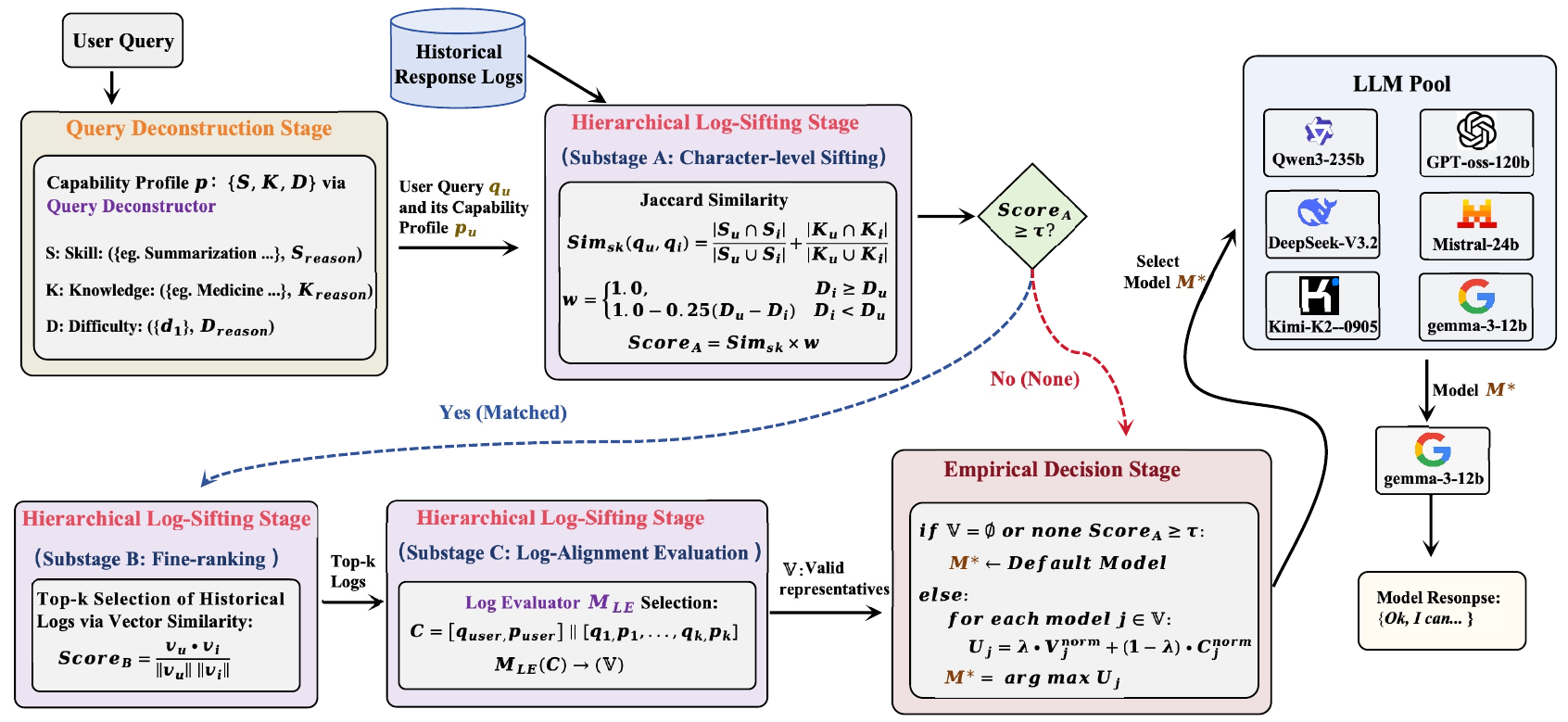}
\caption{Overview of our DecoR framework. The system deconstructs the input query into a capability profile $p=\{S, K, D\}$. It then identifies representative historical logs with aligned capability requirements through a hierarchical three-tier sifting process. Finally, the framework determines the optimal model $M^*$ by balancing performance and cost.}
\vspace{-4pt}
\label{pic:framework}
\end{figure*}
% ----------- pic:verify_in_line  -----------

In this section, we propose DecoR, a model routing architecture designed to derive optimal routing strategies. An overview is illustrated in Figure \ref{pic:framework}. Following the Problem Formulation (Section \ref{method:problem_formulation}), the Query Deconstruction Stage (Section \ref{method:query_deconstruction}) first transforms raw queries into structured capability profiles to capture task-intrinsic requirements. Subsequently, the Hierarchical Log-Sifting Stage (Section \ref{method:log_sifting}) filters raw historical logs to isolate high-value entries. These are then utilized in the Empirical Decision Stage (Section \ref{method:empirical_decision})  to determine the optimal target model by balancing historical performance and operational cost.

\subsection{Problem Formulation}
\label{method:problem_formulation}

Consider a candidate pool of LLMs $\mathcal{M} = \{m_1, \dots, m_T\}$ and a set of historical response logs $\mathcal{H} = \{(q_i, m_{ij}, r_{ij}) : i = 1, \dots, n\}$, where $q_i$ is a historical query, $m_{ij} \in \mathcal{M}$ is the model invoked, and $r_{ij} = (v_{ij}, c_{ij})$ denotes the corresponding execution result, consisting of the performance score $v_{ij}$ and the operational cost $c_{ij}$. In practice, multiple models within $\mathcal{M}$ may adequately satisfy the requirements of a specific query, and not every query necessitates the most powerful yet expensive model. Our objective is to learn a router that selects the most suitable LLM $m^* \in \mathcal{M}$ for each incoming query $q$ by identifying a model that offers sufficient performance while minimizing redundant computational cost.

\subsection{Query Deconstruction Stage}
\label{method:query_deconstruction}

Traditional routing methods~\cite{routellm,routerbench} predominantly operate directly on raw queries. However, proximity in semantic space does not necessitate alignment in capability requirements, rendering systems susceptible to routing deviations caused by superficial linguistic features. To address this, we propose \textbf{Query Deconstruction}, which aims to decouple \textbf{linguistic surface forms} from \textbf{task-intrinsic requirements}. Through this deconstruction mechanism, we shift the routing focus from surface-level textual narratives to deep-seated capability demands, ensuring that the decision-making process is grounded in the essential attributes of the task.

Specifically, we develop a Query Deconstructor $f_{dec}(\cdot)$ that transforms each input query $q$ into a structured \textbf{Capability Profile} $p$:
\begin{equation}
\label{method:formalized1}
p = f_{dec}(q) = \{s, k, d\}
\end{equation}
This profile quantifies three essential dimensions required to fulfill the query:

\begin{itemize}
\item \textbf{Skill Set $(S, s_{reason})$}: $S = \{s_1, s_2, \dots \}$ represents the atomic functional operations (e.g., \textit{information extraction, summarization}) required for $q$. The accompanying $s_{reason}$ provides a concise explanation justifying why these specific skills are necessary.
\item \textbf{Knowledge Domain $(K, k_{reason})$}: $K = \{k_1, k_2, \dots \}$ specifies the domain-specific expertise required (e.g., \textit{medicine, law}). The $k_{reason}$ offers a brief explanation of why these specific knowledge domains are required to address the query. 
\item \textbf{Difficulty $(D, d_{reason})$}: $D \in \{d_0, d_1, d_2, d_3\}$ quantifies the cognitive load. We discretize this complexity into four hierarchical levels, from trivial requests ($d_0$) to deep reasoning tasks ($d_3$). The $d_{reason}$ provides a brief explanation of why the specific difficulty level is assigned to the query.
\end{itemize}
Notably, the specific categories within the Skill Set and Knowledge Domain are not predefined; instead, they are dynamically derived by the Query Deconstructor based on the unique context of each query. The training process for this deconstructor is elaborated in Section \ref{method:training}.

\subsection{Hierarchical Log-Sifting Stage}
\label{method:log_sifting}

This stage aims to precisely extract the most relevant experiences from a massive experience pool to provide a reliable basis for routing. Initially, the system operates offline to augment the Historical Response Logs with capability dimensions using the Query Deconstructor, forming an enhanced library: $\mathcal{H} = \{(q_i, m_{ij}, r_{ij}, p_{i})\}$, where $p_{i} = \{S_{i}, K_{i}, D_{i}\}$. Upon receiving a new query $q_{user}$ and its capability profile $p_{user} = \{S_u, K_u, D_u\}$, the system executes the following three progressive sub-stages.

\subsubsection{Substage A: Character-level Sifting with Capability Constraints} \label{method:stage a}
To ensure retrieval efficiency while enforcing hard capability constraints, the system utilizes an inverted index to match the attributes of $q_u$ against historical logs. We calculate an initial alignment score by independently evaluating the similarity in the skill and knowledge dimensions. Specifically, we employ the Jaccard similarity coefficient to measure the overlap for both Skill Set ($S$) and Knowledge Domain ($K$), and define the lexical similarity $Sim_{sk}$ as their summation:
\begin{equation}Sim_{sk}(q_u, q_i) = \frac{|S_u \cap S_i|}{|S_u \cup S_i|} + \frac{|K_u \cap K_i|}{|K_u \cup K_i|}
\end{equation}
Subsequently, a difficulty matching function $w(D_i, D_u)$ is introduced to calibrate the initial score. If the difficulty of the historical query is not lower than the current query, it is considered a full match; otherwise, the score decreases by 0.25 for each level of deficiency, such that $w(D_i, D_u)$ is defined as:
\begin{equation}
\begin{aligned}
&w=
\begin{cases}
1.0, & D_i \ge D_u \\
1.0 - 0.25 \times (D_u - D_i), & D_i < D_u
\end{cases}
\end{aligned}
\end{equation}
The final sifting score is defined as $Score_A = Sim_{sk} \times w$. The system employs a preset threshold $\tau$ and only retains historical response logs with $Score_A \ge \tau$ for the next stage. If no logs pass this threshold, the user query is identified as out-of-distribution relative to the historical log repository.
Consequently, the system bypasses all subsequent sifting stages and proceeds directly to the Empirical Decision Stage (Section  \ref{method:empirical_decision}).

\subsubsection{Substage B: Fine-ranking}\label{method:stage b}
For the candidate logs passing the Substage A (Section \ref{method:stage a}), the system performs high-dimensional feature matching to capture deep alignments between semantics and capability features. The BGE-M3~\cite{bge-m3} model is used to encode the concatenation of the query text and its capability profile into a feature vector:
\begin{equation}
v = \text{Encoder}(q \oplus \text{String}(p))
\end{equation}
The cosine similarity between the target vector $v_u$ and the candidate vector $v_i$ is then calculated:
\begin{equation}
Score_B(v_u, v_i) = \frac{v_u \cdot v_i}{\|v_u\| \|v_i\|}
\end{equation}
The system ranks the logs in descending order based on $Score_B$ and retains the Top-$k$ most relevant  logs for further processing in Substage C (Section \ref{method:stage c}).

\subsubsection{Substage C: Log-Alignment Evaluation}\label{method:stage c}
As the final phase of the sifting process, the \textbf{Log Evaluator (LE)} filters the refined logs through long-context modeling to extract a subset of truly representative records. The input for LE is a concatenated context $C$ of the current requirements and historical experiences:
\begin{equation}
C = [q_{user}, p_{user}] \parallel [q_1, p_1, \dots, q_k, p_k]
\end{equation}
The LE model $M_{LE}$ generates a reasoning process (Thought) before outputting the set of identifiers $\mathbb{V}$ representing $q_{user}$:
\begin{equation}
\label{method:formalized7}
M_{LE}(C) \rightarrow (\text{Thought}, \mathbb{V})
\end{equation}
where $\mathbb{V}$  is the set of indices for logs identified as valid representatives.  If the LE determines that no logs in the candidate set provide a valid reference, it outputs $\mathbb{V} = \emptyset$. The training methodology for the Query Deconstructor is elaborated in Section \ref{method:training}.
%The resulting set $\mathbb{V}$ serves as the direct evidence for the subsequent empirical decision stage.

\subsection{Empirical Decision Stage}
\label{method:empirical_decision}
Building on the $\mathbb{V} $, this stage determines the final routing decision. A query is categorized as out-of-distribution (OOD) if it falls into either of the following scenarios: (1) it is pre-identified as OOD during the initial sifting in Substage A (Section \ref{method:stage a}), or (2) Substage C (Section \ref{method:stage c}) yields an empty set ($\mathbb{V}  = \emptyset$). 
To ensure robustness, the system invokes an fallback strategy for these OOD queries, rerouting them to a high-performance model to guarantee high-quality responses.

If $\mathbb{V}  \neq \emptyset$, the decision-maker initiates an empirical inference procedure. To address the magnitude difference between performance scores $v_{ij}$ and inference costs $c_{ij}$, a normalization procedure is applied to balance their relative influence. Specifically, the system performs empirical data aggregation by retrieving the ground-truth performance of candidate model $j$ from the historical library $\mathcal{H}$ for tasks corresponding to indices in $\mathbb{V} $. The average performance $\bar{V}_j$ and average cost $\bar{C}_j$ for each model are calculated as:
\begin{equation}
\bar{V}_j = \frac{1}{|\mathbb{V}|} \sum_{i \in \mathbb{V}} v_{ij}, \quad \bar{C}_j = \frac{1}{|\mathbb{V}|} \sum_{i \in \mathbb{V}} c_{ij}
\end{equation}

Subsequently, the system performs dual-dimensional dimensionless processing via linear normalization. This process transforms absolute values into relative scores within the $[0, 1]$ interval. The performance utility score $V^{norm}_j$ and cost utility score $C^{norm}_j$ are defined as:
\begin{equation}
\begin{aligned}
V^{norm}_j = \frac{\bar{V}_j - \min(\bar{V})}{\max(\bar{V}) - \min(\bar{V}) + \epsilon} \\
C^{norm}_j = \frac{\max(\bar{C}) - \bar{C}_j}{\max(\bar{C}) - \min(\bar{C}) + \epsilon}
\end{aligned}
\end{equation}
where $\epsilon$ is an infinitesimal constant to prevent division by zero. This transformation provides a balanced quantitative representation of performance and cost, eliminating the influence of disparate scales.

Finally, a balancing factor $\lambda \in [0, 1]$ is introduced to adjust the preference between performance and economy. The comprehensive utility score $U_j$ for each model is computed as:
\begin{equation}
U_j = \lambda \cdot V^{norm}_j + (1 - \lambda) \cdot C^{norm}_j
\end{equation}
The routing decision-maker selects the model with the highest utility score as the optimal target: 
\begin{equation}
m^* = \text{arg}\max_j U_j
\end{equation}
where $m^*$ denotes the final model selected to generate the response for the user's query.

\subsection{Training}
\label{method:training}

\subsubsection{Query Deconstructor}
\label{method:train_deconstructor}
The primary objective of the Query Deconstructor is to decompose a raw query $q$ into a structured capability profile $p$, as formalized in Eq. (\ref{method:formalized1}). Training data are synthesized via GPT-5 and further refined through a rigorous expert review process involving three CS PhD students (detailed protocols and prompts are provided in Appendix \ref{apex:deconstructor_details}). Through this high-quality instruction tuning, the model learns to internalize complex task decomposition patterns. The module is then trained via Supervised Fine-Tuning (SFT) using the following loss function:
\begin{equation} 
\mathcal{L}_{dec} = - \sum_{t=1}^{T} \log P(y_t | y_{<t}, q)
\end{equation}
where $q$ represents the original input query and $y$ denotes the expert-validated structured Capability Profile $p$.

\subsubsection{Log Evaluator}
\label{method:train_evaluator}
The Log Evaluator implements the mapping defined in Eq. (\ref{method:formalized7}) to select a representative log set $\mathbb{V}$ from the input context. To foster autonomous judgment and ensure robust generalization in OOD scenarios, the module is optimized via Group Relative Policy Optimization (GRPO~\cite{deepseek-math}). 
To quantify the alignment between the predicted validation set $\mathbb{V}$ and the ground truth $G$, we define the reward function $R(\mathbb{V}, G)$ as:

{
\small
\begin{equation}
\left\{
\begin{aligned}
&6, && \text{if } \mathbb{V} = G \\
&-2 |\mathbb{V}|, && \text{if } G = \emptyset \land \mathbb{V} \neq \emptyset \\
&-6, && \text{if } G \neq \emptyset \land \mathbb{V} \cap G = \emptyset \\
&\textstyle \frac{6}{|G|} (|\mathbb{V} \cap G| - |\mathbb{V} \setminus G|), && \text{otherwise}
\end{aligned}
\right.
\end{equation}
}
where $|\mathbb{V} \cap G|$ and $|\mathbb{V} \setminus G|$ denote the number of hits and false positives, respectively. This formulation incentivizes high recall while penalizing hallucinations and retrieval failures.
The GRPO training paradigm enables the model to learn the intrinsic utility logic of historical logs rather than relying on rigid classification patterns. 
The optimization objective is defined as:

{
\small
\begin{equation}
\mathcal{J}_{\text{GRPO}}(\theta) = \mathbb{E}_{\genfrac{}{}{0pt}{2}{(q,a) \sim \mathcal{D}}{o_i \sim \pi_{{\text{old}}}}}
\left[
\frac{1}{\sum_{i=1}^G |o_i|} \sum_{i=1}^G \sum_{t=1}^{|o_i|} L_{i,t}
\right]
\end{equation}
}

where $o_i$ denotes the $i$-th sampled output for a given query $q$; $G$ represents the number of sampled outputs per query, $L_{i,t}$ denotes the loss function:

{
\small
\begin{equation}
L_{i,t} = \min( r_{i,t} \hat{A}_{i,t},\ \text{clip}\left(r_{i,t},\ 1-\varepsilon,\ 1+\varepsilon\right) \hat{A}_{i,t} ),
\end{equation}
}

where $r_{i,t}$ is the importance weight and $\hat{A}_{i,t}$ denotes the normalized advantage.
In the following, we describe the training data construction process for the Log Evaluator.

Further details regarding the training paradigm and data construction methodology are provided in Appendix \ref{apex:evaluator_details}.

% \section{Experiments Setup}
\section{Experiments}

\subsection{Experiments Setup}
\subsubsection{Datasets and Metrics}
\textbf{Datasets} \ 
% In this study, we construct an experimental benchmark named \textbf{CodaSet}, which comprises two primary components: In-distribution (ID) tasks and OOD evaluations. The ID tasks encompass MMLU-Pro~\cite{mmlu_pro}, GSM8K~\cite{gsm8k}, IFEval~\cite{ifeval}, and BBH~\cite{bbh}. For each task type, the corresponding datasets are partitioned into training and test sets; notably, both our proposed model and all baseline models are trained exclusively on these ID training sets. The OOD evaluation suite consists of Math500~\cite{math500}, MT-bench~\cite{mt_bench}, and MBPP~\cite{mbpp}. These datasets are used only during the testing phase to evaluate the model’s generalization capability across unseen tasks. For further details regarding the construction of CodaSet and statistical distributions, please refer to Appendix \ref{apex:dataset}.
We construct CodaSet, comprising ID tasks (MMLU-Pro~\cite{mmlu_pro}, GSM8K~\cite{gsm8k}, IFEval~\cite{ifeval}, and BBH~\cite{bbh}) and OOD evaluations (Math500~\cite{math500}, MT-bench~\cite{mt_bench}, and MBPP~\cite{mbpp}). Both the model training and our historical log corpus rely exclusively on the ID training sets. During evaluation, the test set encompasses the ID test partitions alongside the complete OOD datasets to assess both task-specific performance and generalization capability. Further details are provided in Appendix \ref{apex:dataset}.

\noindent \textbf{Evaluation Metrics} \ 
% We follow standardized evaluation protocols established in prior research for each task. For tasks including MMLU-Pro, GSM8K, Math500, and BBH, we adopt the Exact Match principle, calculating accuracy by comparing the model-generated answers with the ground truth. For IFEval and MBPP, we utilize the evaluation scripts provided by their official repositories for automated assessment. Regarding MT-bench, we employ the LLM-as-a-Judge framework to score the quality of the model-generated responses. Detailed evaluation metrics for each dataset are provided in Appendix \ref{apex:metrics}.
% We follow standard protocols: Exact Match (EM) accuracy for MMLU-Pro, GSM8K, Math500, and BBH, and official evaluation scripts for IFEval, MBPP, and MT-bench. Detailed metrics are provided in Appendix \ref{apex:metrics}.
We employ a diverse set of metrics tailored to each task. For MMLU-Pro, GSM8K and BBH, we use Exact Match (EM) by extracting final answers via regular expressions for ground-truth comparison to calculate accuracy. For specific domains, we leverage established evaluation frameworks: Math500 is assessed using OpenAI's \texttt{simple-evals} framework\footnote{\url{https://github.com/openai/simple-evals}}, MBPP is evaluated via the EvalPlus\footnote{\url{https://github.com/evalplus/evalplus}} to ensure rigorous code correctness. For MT-Bench, we adopt the FastChat LLM-as-a-Judge\footnote{\url{https://github.com/lm-sys/FastChat/tree/main/fastchat/llm_judge}} evaluation protocol and employ GPT-5.1 as the judge to score responses on a 0.1–1 scale. 
For IFEval, we utilize its official automated script\footnote{\url{https://github.com/google-research/google-research/tree/master/instruction_following_eval}} to verify strict constraint adherence. To ensure stability, accuracy-based metrics are determined via majority voting across five independent runs, while MT-bench scores are reported as the arithmetic mean of these iterations.

\subsubsection{Implementation Details}
We employ \textbf{Qwen3-0.6B}\footnote{\url{https://huggingface.co/Qwen/Qwen3-0.6B}} as the base model for both the Deconstructor and the Log Evaluator. During the training phase, the Deconstructor is optimized via Supervised Fine-Tuning with a learning rate of $2 \times 10^{-5}$. The \textbf{Log Evaluator} is trained using the VERL\footnote{\url{https://github.com/volcengine/verl}} reinforcement learning framework with a learning rate of $1 \times 10^{-6}$. Detailed hyperparameter configurations for model training are provided in Appendix \ref{appendix:hyperparameters}.
% We empirically set $\tau = 0.5$ for Substage A and $k = 3$ for Substage B, following tuning on a small-scale validation set.
% Additionally, in the Empirical Decision Stage, the weighting parameter $\lambda$ is set to 0.5. 
Based on validation set tuning, we set $\tau = 0.5$, $k = 3$, and $\lambda = 0.5$.
All results are averaged over three independent trials.

% We adopt \textbf{Qwen3-0.6B}\footnote{\url{https://huggingface.co/Qwen/Qwen3-0.6B}} for both Deconstructor (SFT, LR $2 \times 10^{-5}$) and Log Evaluator (trained via verl\footnote{\url{https://github.com/volcengine/verl}} RL, LR $1 \times 10^{-6}$). Based on validation set tuning, we set $\tau = 0.5$, $k = 3$, and $\lambda = 0.5$. Detailed configurations are provided in Appendix \ref{appendix:hyperparameters}. All results are averaged over three independent trials.

\subsubsection{Comparison Methods and LLM Pool}
\textbf{Comparison Methods} \ 
To ensure a comprehensive evaluation, we compare DecoR with several representative baselines:
(1) Random Router~\cite{routellm}, (2) LLM Router, (3) RouterDC~\cite{RouterDC}, (4) KNN Router~\cite{routerbench}, (5) EmbedLLM~\cite{embedllm}, (6) MODEL-SAT~\cite{SAT}.
% (5) EmbedLLM~\cite{embedllm}, 
All baseline routers are trained using the training split of CodaSet to ensure a fair comparison.
Detailed descriptions of these baselines are provided in Appendix \ref{apex:detailed_comparison_methods}.

\noindent \textbf{LLM Pool} \ 
% We construct a diverse LLM pool including:
We construct a diverse LLM pool consisting of eight representative models with varying sizes, including Kimi-K2-Instruct-0905~\cite{kimi_k2}, DeepSeek-V3.1-Terminus~\cite{deepseekv31}, DeepSeek-V3.2-Exp~\cite{deepseekv32}, Qwen3-235B-A22B-Instruct~\cite{qwen3_235}, gpt-oss-120b~\cite{gptoss120},  gemma-3-27b-it~\cite{gemma3}, Mistral-Small-3.2-24B-Instruct and gemma-3-12b-it~\cite{gemma3}.
All candidate models are accessed through the DeepInfra API\footnote{\url{deepinfra.com}} to retrieve inference results and collect empirical cost data under real-world deployment settings.
Please refer to Appendix \ref{apx:base_models} for comprehensive descriptions of these models.

% Our model pool includes eight diverse LLMs: Kimi-K2-0905, DeepSeek-V3.1-Terminus, DeepSeek-V3.2-Exp, Qwen3-235B, gpt-oss-120b, gemma-3 (12B/27B), and Mistral-Small-24B.  All candidate models are accessed via the DeepInfra API\footnote{\url{deepinfra.com}} to ensure real-world inference and cost evaluations. Please refer to Appendix \ref{apx:base_models} for full model descriptions.

% \section{Experimental Analysis}

\subsection{Main Results}

\begin{table*}[ht]
\centering
\resizebox{\textwidth}{!}{
\begin{tabular}{l|cc|cc|cc|cc|cc}
\toprule
\textbf{Model} & \multicolumn{2}{c|}{\textbf{GSM8k}} & \multicolumn{2}{c|}{\textbf{MMLU-PRO}} & \multicolumn{2}{c|}{\textbf{BBH}} & \multicolumn{2}{c|}{\textbf{IFEVAL}} & \multicolumn{2}{c}{\textbf{Average}} \\
 & Perf (\%) & Cost & Perf (\%) & Cost & Perf (\%) & Cost & Perf (\%) & Cost & Perf (\%) & Cost \\
\midrule
\multicolumn{11}{c}{\textit{LLM Pool}} \\
\midrule
Kimi-K2-Instruct & 94.30 & 11.1x & 81.75 & 11.6x & 90.54 & 13.7x & 85.17 & 12.8x & 87.94 & 11.9x \\

DeepSeek-V3.1 & 94.00 & 5.1x & 83.94 & 5.2x & 92.27 & 5.8x & \textbf{88.76} & 4.9x & 89.74 & 5.2x \\

DeepSeek-V3.2-Exp & 92.20 & 4.3x & 83.76 & 4.0x & 91.46 & 4.2x & 86.60 & 2.4x & 88.51 & 3.9x \\

Qwen3-235B-A22B & \textbf{96.40} & 3.8x & \textbf{84.86} & 5.8x & \textbf{94.29} & 5.4x & 86.84 & 3.3x & \textbf{90.60} & 5.0x \\

gpt-oss-120b & 94.20 & 2.4x & 80.21 & 2.8x & 92.85 & 2.7x & 87.56 & 3.1x & 88.71 & 2.7x \\

gemma-3-27b-it & 92.80 & 1.6x & 71.58 & 3.5x & 86.10 & 1.5x & 84.69 & 1.4x & 83.79 & 2.4x \\

Mistral-Small-3.2-24B & 53.10 & 2.0x & 73.77 & 1.9x & 85.64 & 1.7x & 77.03 & \textbf{1.0x} & 72.39 & 1.8x \\

gemma-3-12b-it & 94.50 & \textbf{1.0x} & 69.64 & \textbf{1.0x} & 85.35 & \textbf{1.0x} & 81.82 & 1.2x & 82.83 & \textbf{1.0x} \\

\midrule
\multicolumn{11}{c}{\textit{Router Baselines}} \\
\midrule
Random Router & 87.83 & 4.1x & 77.92 & 4.8x & 88.06 & 4.7x & 82.93 & {3.0x} & 84.18 & 4.4x \\
LLM Router & 94.27 & 4.3x & 65.56 & {1.1x} & 84.90 & {1.1x} & 84.19 & 3.6x & 82.23 & 2.3x \\
KNN Router & 93.93 & {1.2x} & 77.51 & 2.5x & 92.16 & 1.6x & 82.78 & 3.3x & 86.60 & 2.0x \\
% SVM Router &90.66	&5.2x	&77.29	&3.5x	&92.16	&5.8x	&85.41	&3.2x	&86.38	&4.4x \\
RouterDC & 90.10 & 3.8x & 79.99 & 4.4x & 91.58 & 4.4x & 85.41 & 5.0x & 86.77 & 4.2x \\
EmbedLLM& 87.82  & 4.2x & 79.74 & 4.3x & 92.79 & 4.7x & 86.84 & 3.2x & 86.80 & 4.2x \\
MODEL-SAT & 90.60 & 5.2x & \colorbox{pink!77}{81.82} & 6.0x & 92.16 & 5.8x & 85.41 & 3.5x & 87.50 & 7.8x \\
% EmbedLLM& 90.60 & 5.2x & \colorbox{pink!77}{81.82} & 6.0x & 92.16 & 5.8x & 85.41 & 3.5x & 87.50 & 7.8x \\
% EmbedLLM& 96.13 & 3.9x & 84.22 & 5.1x & 94.50 & 4.5x & 82.90 & 0.9x \\

\midrule
\multicolumn{11}{c}{\textit{Proposed Methods}} \\
\midrule
DecoR (DeepSeek-V3.1) & \colorbox{pink!77}{95.59} & 1.4x & 80.36 & 2.2x & \colorbox{pink!77}{93.89} & 1.9x & \colorbox{pink!77}{87.56} & 4.5x & \colorbox{pink!77}{89.35} & 2.1x \\
DecoR (gpt-oss-120b) & 95.54 & 1.3x & 79.77 & 2.0x & \colorbox{pink!77}{93.89} & 1.9x & 87.32 & {3.0x} & 89.13 & {1.9x} \\
\bottomrule
\end{tabular}
}
\caption{Performance and cost comparison on the CodaSet ID test set. Performance is quantified as Accuracy (Acc), with raw scores scaled by 100 (\%) for clarity. Within the LLM Pool, \textbf{bold} values indicate the best performance. For Router Baselines and Proposed Methods, \colorbox{pink!77}{pink}  represent the highest performance across these two groups, respectively. \textbf{Higher performance and lower cost are more desirable.} The notation DecoR ($\cdot$) signifies the specific model employed as the fallback model. }
\label{exp:main_id}
\end{table*}

Tables \ref{exp:main_id} and \ref{tab:ood_results} present the experimental results in both ID and OOD scenarios. In ID settings, our proposed DecoR (DeepSeek-V3.1) significantly outperforms the majority of router baselines while maintaining much lower computational overhead. Specifically, although MODEL-SAT achieves a slightly higher score than DecoR on MMLU-PRO, its computational cost is three times as high. On average, DecoR’s performance approximates that of the strongest single model, Qwen3-235B-A22B, and even surpasses it on the IFEVAL benchmark, despite Qwen3's cost being 2.4 times that of our method (5.0x vs. 2.1x).
Furthermore, DecoR demonstrates remarkable robustness in OOD scenarios. As shown in Table \ref{tab:ood_results}, DecoR experiences only a marginal performance decline in OOD settings, whereas other baseline methods suffer from significant degradation, with some even performing worse than the Random Router. This stability stems from our system's adaptive mechanism: when encountering OOD queries where prior experience is insufficient, the system triggers a fallback logic to a pre-specified high-performance model (DeepSeek-V3.1) rather than making erroneous assignments as the baselines do. Although this strategy leads to a localized increase in cost, it effectively preserves performance stability in unknown domains. In terms of average performance, DecoR remains competitive with top-tier single models while retaining a clear cost advantage.

\begin{table}[htbp]
\centering
\small
\resizebox{\linewidth}{!}{
\begin{tabular}{lcccccc}
\toprule
\multirow{2}{*}{\textbf{Method}} & \multicolumn{2}{c}{\textbf{GSM8k}} & \multicolumn{2}{c}{\textbf{MMLU-PRO}} & \multicolumn{2}{c}{\textbf{BBH}} \\ \cmidrule(lr){2-3} \cmidrule(lr){4-5} \cmidrule(lr){6-7} 
 & Perf. & Cost & Perf. & Cost & Perf. & Cost \\ \midrule
DecoR (Full) & \textbf{95.54} & 1.1$\times$ & \textbf{79.77} & 1.0$\times$ & \textbf{93.89} & 1.3$\times$ \\
w/o Satge A & 91.70 & 2.1$\times$ & 76.16 & 1.8$\times$ & 88.57 & 1.8$\times$ \\
w/o Satge B & 94.30 & 1.0$\times$ & 77.29 & 1.1$\times$ & 90.47 & 1.0$\times$ \\
w/o Satge C & 94.32 & 1.0$\times$ & 78.17 & 1.0$\times$ & 91.25 & 1.1$\times$ \\ \bottomrule
\end{tabular}
}
\caption{Ablation study of DecoR components on three benchmarks. The best results are \textbf{bolded}. Cost is normalized by the column-wise minimum to indicate the relative computational overhead ($\times$).}
\label{tab:ablation_study}
\end{table}

\begin{table*}[htbp]
\centering
\resizebox{0.87\textwidth}{!}{
\begin{tabular}{l|cc|cc|cc|cc}
\toprule
\textbf{Model} & \multicolumn{2}{c|}{\textbf{Math\_500}} & \multicolumn{2}{c|}{\textbf{MBPP}} & \multicolumn{2}{c|}{\textbf{MT\_Bench}} & \multicolumn{2}{c}{\textbf{Average}} \\
 & Perf (\%) & Cost & Perf (\%) & Cost & Perf (\%) & Cost & Perf (\%) & Cost \\
\midrule
\multicolumn{9}{c}{\textit{LLM Pool}} \\
\midrule
Kimi-K2-Instruct & 81.31 & 15.2x & 77.78 & 20.1x & 94.47 & 7.9x & 84.52 & 14.9x \\
DeepSeek-V3.1 & 89.86 & 5.6x & 75.66 & 14.1x & 97.24 & 5.3x & 87.59 & 7.1x \\
DeepSeek-V3.2-Exp & \textbf{90.33} & 3.5x & 75.13 & 12.2x & 95.53 & 2.4x & 87.00 & 6.8x \\
Qwen3-235B & 85.98 & 5.3x & \textbf{79.37} & 13.8x & \textbf{98.16} & 3.1x & \textbf{87.84} & 6.4x \\
gpt-oss-120b & 83.18 & 1.9x & 73.54 & 11.6x & \textbf{98.16} & 2.9x & 84.96 & 3.8x \\
gemma-3-27b-it & 78.04 & 1.3x & 76.19 & 1.3x & 95.53 & 1.2x & 83.25 & 1.3x \\
Mistral-3.2-24B & 76.64 & 1.4x & 73.02 & 6.0x & 93.68 & 1.2x & 81.11 & 2.2x \\
gemma-3-12b-it & 75.70 & \textbf{1.0x} & 76.19 & \textbf{1.0x} & 92.37 & \textbf{1.0x} & 81.42 & \textbf{1.0x} \\
\midrule
\multicolumn{9}{c}{\textit{Router Baselines}} \\
\midrule
Random Router & 80.18 & 4.0x & 71.96 & 8.9x & 95.26 & 3.7x & 82.47 & 4.9x \\
LLM Router & 77.57 & 1.2x & 73.19 & 4.0x & 94.34 & 1.7x & 81.70 & 1.8x \\
KNN Router & 77.10 & {1.1x} & 73.54 & {2.7x} & 93.95 & {1.4x} & 81.53 & {1.4x} \\
RouterDC & 80.64 & 3.9x & 72.66 & 6.9x & 95.00 & 2.4x & 82.77 & 4.2x \\
% SVM Router &79.98	&2.2x	&71.60	&4.7x	&94.21	&2.0x	&81.93 &3.0x \\
EmbedLLM & 82.71 & 5.1x & 73.72 & 8.4x & 94.74 & 3.2x & 83.72 & 5.4x \\
MODEL-SAT & 81.98 & 2.9x & 74.60 & 4.7x & 94.21 & 2.0x & 83.60 & 3.3x \\
% EmbedLLM & 82.71 & 5.1x & 73.72 & 8.4x & 94.74 & 3.2x & 83.72 & 5.4x \\

\midrule
\multicolumn{9}{c}{\textit{Proposed Methods}} \\
\midrule
DecoR (DeepSeek-V3.1) & \colorbox{pink!77}{85.51} & 3.1x & \colorbox{pink!77}{76.72} & 12.4x & 97.24 & 4.7x & \colorbox{pink!77}{86.49} & 5.0x \\
DecoR (gpt-oss-120b) & 80.37 & 1.4x & 72.49 & 10.2x & \colorbox{pink!77}{98.16} & 2.7x & 83.67 & 3.3x \\
\bottomrule
\end{tabular}
}
\caption{Performance and cost comparison of the LLM pool and routing baselines on the CodaSet OOD test set. }
\label{tab:ood_results}
\end{table*}
% SVM Router &85.98	&3.2x	&61.86 &4.7x	&74.60	&4.7x	&94.21	&2.0x	&79.16 &3.7x

\subsection{Ablation Study}
The ablation results in Table \ref{tab:ablation_study} demonstrate that the full DecoR system achieves the best performance across all benchmarks. Removing Stage A (Query Deconstruction) causes the most significant decline in accuracy and a sharp cost increase of up to 2.1$\times$ on GSM8k, proving that task decomposition is essential for simplifying the reasoning space and reducing redundant computations. While removing Stage B (Fine-ranking) or Stage C (Log-Alignment Evaluation) also leads to noticeable performance drops, their impact on computational overhead is minimal. This confirms the necessity and collective contribution of each module within our proposed framework.

\subsection{Analysis}

\noindent \textbf{Impact of Base Model Scale} \
The scaling analysis of the Qwen3 backbone reveals that while performance generally improves with model size, the gains are marginal compared to the increased computational overhead. As shown in Table \ref{tab:qwen3_scaling_results}, the 4b model achieves only a slight average improvement over the 0.6b variant, specifically 0.11\% in ID scenarios and 0.89\% in OOD scenarios. Consequently, we select the 0.6b model as our primary backbone to keep the framework lightweight and cost-effective for practical deployment.

\begin{table}[htbp]
\centering
\small
\resizebox{\linewidth}{!}{
\begin{tabular}{lcccccccc}
\toprule
\multirow{2}{*}{\textbf{Size}} & \multicolumn{2}{c}{\textbf{GSM8k}} & \multicolumn{2}{c}{\textbf{MMLU-PRO}} & \multicolumn{2}{c}{\textbf{BBH}} & \multicolumn{2}{c}{\textbf{ID Avg}} \\ \cmidrule(lr){2-3} \cmidrule(lr){4-5} \cmidrule(lr){6-7} \cmidrule(lr){8-9}
 & Perf. & Cost & Perf. & Cost & Perf. & Cost & Perf. & Cost \\ \midrule
0.6b & 95.59 & 1.0$\times$ & 80.36 & 1.0$\times$ & 93.89 & 1.0$\times$ & 89.95 & 1.0$\times$ \\
1.7b & 95.59 & 1.0$\times$ & 80.42 & 1.0$\times$ & 93.89 & 1.0$\times$ & 89.97 & 1.0$\times$ \\
4.0b   & \textbf{95.63} & 1.0$\times$ & \textbf{80.63} & 1.0$\times$ & \textbf{93.92} & 1.0$\times$ & \textbf{90.06} & 1.0$\times$ \\ \midrule
\midrule
\multirow{2}{*}{\textbf{Size}} & \multicolumn{2}{c}{\textbf{Math\_500}} & \multicolumn{2}{c}{\textbf{MT\_Bench}} & \multicolumn{2}{c}{\textbf{MBPP}} & \multicolumn{2}{c}{\textbf{OOD Avg}} \\ \cmidrule(lr){2-3} \cmidrule(lr){4-5} \cmidrule(lr){6-7} \cmidrule(lr){8-9}
 & Perf. & Cost & Perf. & Cost & Perf. & Cost & Perf. & Cost \\ \midrule
0.6b & 85.51 & 1.0$\times$ & 98.16 & 1.0$\times$ & 76.72 & 1.1$\times$ & 86.79 & 1.0$\times$ \\
1.7b & 85.51 & 1.0$\times$ & 98.16 & 1.0$\times$ & \textbf{77.25} & 1.0$\times$ & 86.97 & 1.0$\times$ \\
4.0b   & \textbf{85.98} & 1.1$\times$ & \textbf{98.27} & 1.1$\times$ & 77.19 & 1.1$\times$ & \textbf{87.15} & 1.1$\times$ \\ \bottomrule
\end{tabular}
}
\caption{Performance and cost comparison using different sizes of Qwen3 models as the base for Deconstructor and Evaluator. The best performance in each column is \textbf{bolded}. Cost is normalized by the column-wise minimum to indicate relative overhead ($\times$). }
\label{tab:qwen3_scaling_results}
\end{table}

\begin{figure}[!t]
\centering
\includegraphics[width=0.99\linewidth]{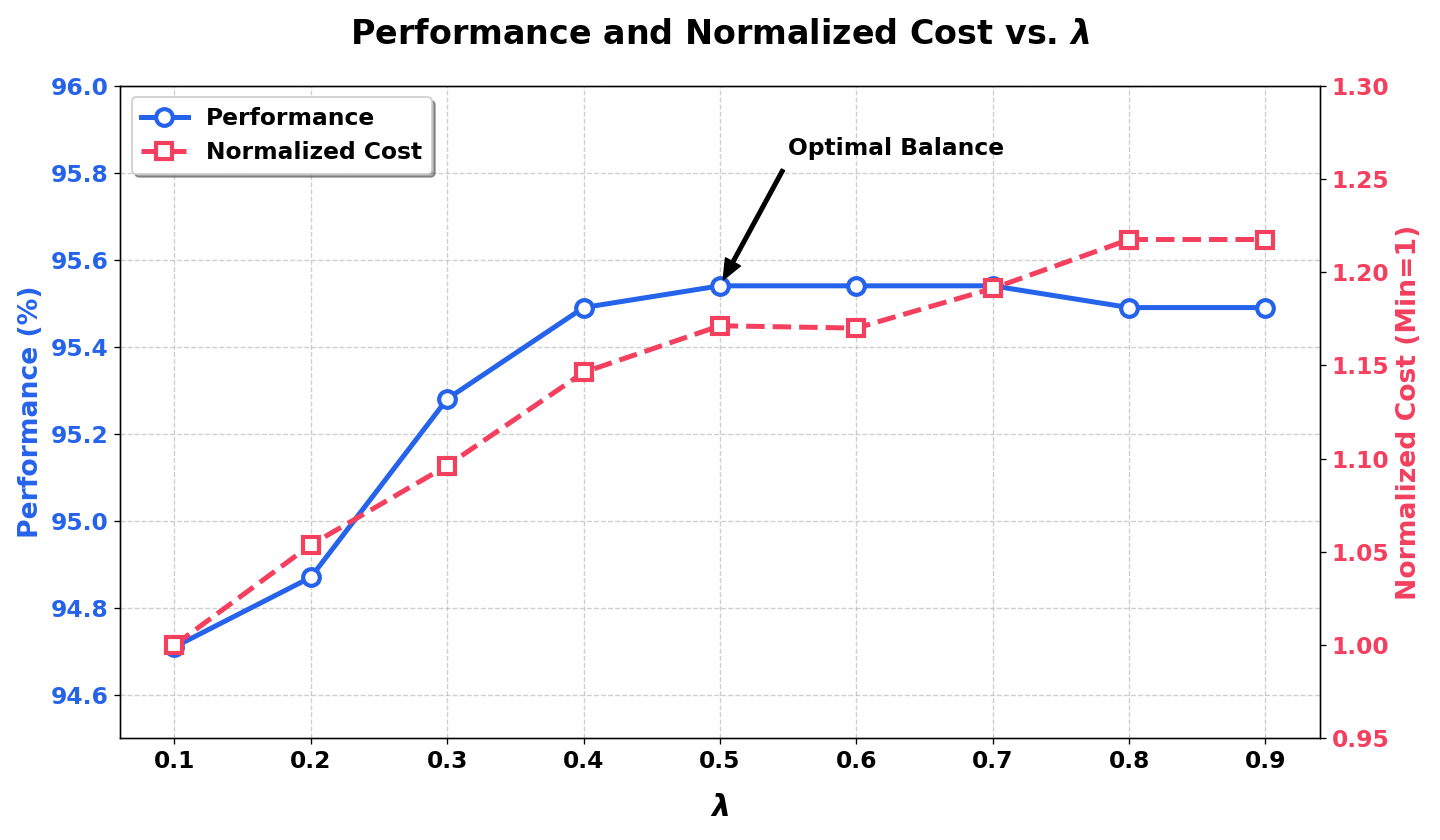}
\caption{Performance and normalized cost vs. trade-off parameter $\lambda$ on the GSM8k dataset.}
\vspace{-4pt}
\label{pic:lamda}
\end{figure}

% \subsection{Impact of \texorpdfstring{$\lambda$}{lambda} on Decision Utility}
% τ λ topk 得选择  stage 拒绝掉的个数
\noindent \textbf{Impact of \texorpdfstring{$\lambda$}{lambda} on Decision Utility} \
We evaluate the sensitivity of $\lambda$, which weights the system's preference for performance over cost on the GSM8k dataset. As illustrated in Figure \ref{pic:lamda}, performance improves substantially as $\lambda$ increases toward 0.5. At $\lambda = 0.5$, the system achieves an Optimal Balance, maximizing the margin between performance gain and normalized cost. Beyond this threshold ($\lambda > 0.5$), performance plateaus as candidate models reach their inherent capacity; however, the cost escalates sharply as the router increasingly selects expensive models for marginal accuracy gains.
% ----------- pic:verify_in_line  -----------
% ----------- pic:verify_in_line  -----------

\begin{figure}[!t]
\centering
\includegraphics[width=0.99\linewidth]{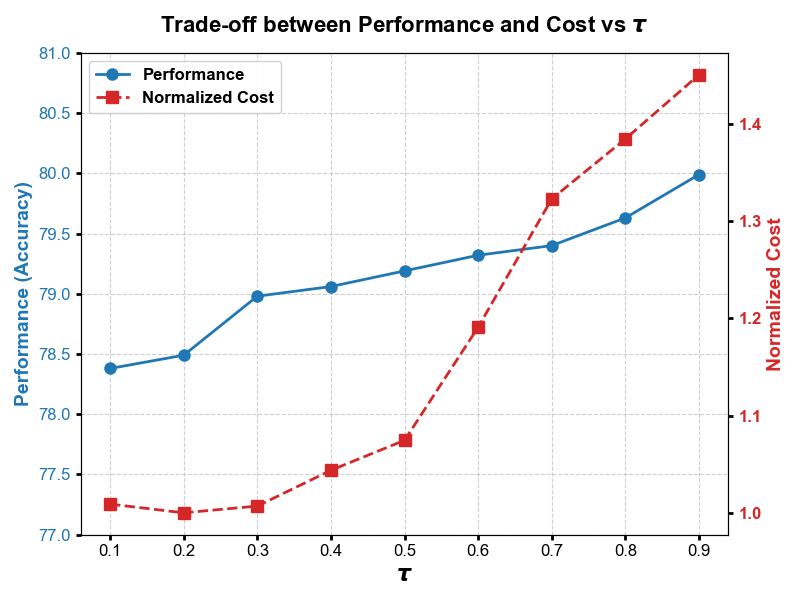}
\caption{Trade-off between performance and cost across varying $\tau$ values. The figure illustrates the trends for model accuracy (solid blue line) and normalized cost (dashed red line) as $\tau$ ranges from 0.1 to 0.9. As $\tau$ increases, model performance exhibits an upward trend, accompanied by a corresponding increase in computational cost.}
\vspace{-4pt}
\label{pic:tau}
\end{figure}

\noindent \textbf{Impact of Shifting Threshold $\tau$ in Substage A} \
% We explore the sensitivity of Substage A to the threshold $\tau$ within the range $[0.1, 0.9]$. A more comprehensive analysis of this impact is provided in Appendix \ref{apex:thresh_tau}.
We explore the sensitivity of Substage A to the threshold $\tau$ within the range $[0.1, 0.9]$. As illustrated in Figure \ref{pic:tau}, increasing $\tau$ leads to a gradual improvement in model performance, alongside a corresponding rise in computational cost. In the low $\tau$ regime, performance is relatively limited, though costs remain minimal. As $\tau$ enters the mid-range, performance gains become significant while cost growth remains moderate. However, in the high $\tau$ range, performance tends to saturate while cost increases become more pronounced. Balancing performance enhancement and cost control, we select $\tau = 0.5$ as it achieves an optimal trade-off and serves as our default configuration.

\subsection{Case Study}
% \noindent \textbf{Case study} \
To provide a clearer understanding of the internal decision-making logic of DecoR, we conduct case study experiments. A more comprehensive analysis is provided in Appendix \ref{apex:case_study}.

\section{Conclusion}
In this paper, we present DecoR, a routing framework that recasts routing as a log-matching process to effectively mitigate the memorization trap. By decoupling task requirements from surface forms, DecoR grounds decisions in capability dimensions, enhancing both accuracy and robustness. To allow for rigorous evaluation, we introduce the CodaSet benchmark, where DecoR demonstrates superior accuracy and cost-efficiency across both ID and OOD settings. Beyond performance, DecoR enables sustainable system evolution by allowing for seamless iteration via log updates without model retraining.

\section*{Limitations}
Although DecoR significantly outperforms existing routing baselines in both ID and OOD scenarios while achieving competitive performance at lower costs and offering seamless extensibility through log updates, several areas for optimization remain.
First, the scoring process during the construction of historical logs is influenced to some extent by the performance of LLM-as-a-judge. This connection implies that there is still potential for growth in achieving fully automated online updates, where new data could be synchronized into the historical repository in real time during the inference process to continuously enhance the system's knowledge base. We intend to consistently refine this feature as the evaluation capabilities of large language models evolve.
Second, the current system architecture could be further improved by incorporating a filtering mechanism for inputs that are highly similar to existing queries in the historical repository to avoid data redundancy. In future research, we will work on developing efficient deduplication and filtering functions to enhance system efficiency and provide more robust support for the research community.

\bibliography{custom}

\appendix

% \section{Example Appendix}
% \label{sec:appendix}

\section{Data Synthesis and Expert Review Protocol} \label{data_human}
\subsection{Query Deconstructor}\label{apex:deconstructor_details}
\textbf{Data Selection and Labeling} \ 
To construct a high-quality dataset for query decomposition, we leveraged GPT-5.1\footnote{\url{https://platform.openai.com/docs/models/gpt-5.1}} to synthesize initial query-profile pairs. This process employed a multi-stage prompting strategy as detailed in Table \ref{table:capability-prompt}. The strategy guided the model to identify core intents, decompose necessary capabilities, and format the output into structured Capability Profiles. This approach ensures that the synthetic data captures the nuanced requirements of complex user queries while maintaining a consistent format for supervised fine-tuning.

\noindent \textbf{Expert Verification Process} \
To ensure the logical integrity of the synthesized samples, we implemented a stringent human-in-the-loop verification process. Three PhD students specializing in Computer Science independently audited each sample based on the criteria of correctness, granularity, and completeness. We adopted a unanimous consensus rule where a sample was incorporated into the final high-fidelity dataset only if it received a ``Pass'' score from all three experts. This collaborative filtering mechanism effectively eliminated hallucinations and logical inconsistencies, resulting in a reliable training set for the Query Deconstructor.

\begin{table*}[ht]
\centering
\small
\renewcommand{\arraystretch}{1.5} % 增加行间距，1.5表示为原来的1.5倍
\begin{tabularx}{\textwidth}{lX}
\toprule
 & \textbf{System Prompt:} \\
 & You are a Capability Decomposition Engine. Your task is to decompose the user query into its capability-space representation $C(q) = \{S, K, D\}$. Follow all rules strictly and output JSON only. \\
 \midrule
 & \textbf{Instruction:} \\
 & \$\{query\} \\
 % \midrule
\textbf{Template} & \textbf{Decomposition Rules:} \\
 & 1. \textbf{Skill Set (S):} Identify required skills (e.g., reasoning, coding). Output as a list and provide "S\_reason". \\
 & 2. \textbf{Knowledge Domain (K):} Identify domains (e.g., law, finance). If none, output "none". Output as a list and provide "K\_reason". \\
 & 3. \textbf{Difficulty (D):} Choose exactly one from \{D0, D1, D2, D3\} based on complexity. Provide "D\_reason". \\
 & \\
 & \textbf{Important Rules:} \\
 & - Output MUST be valid pure JSON. \\
 & - Do NOT include markdown code. \\
 % \midrule
 & \textbf{Output Format (STRICT):} \\
 & \{ \\
 & \quad "S": [...], "S\_reason": "...", \\
 & \quad "K": [...], "K\_reason": "...", \\
 & \quad "D": "...", "D\_reason": "..." \\
 & \} \\
\bottomrule
\end{tabularx}
\caption{The template used for the Capability Decomposition Engine. The Query Deconstructor also utilizes this exact prompt template for both its training and inference phases}
\label{table:capability-prompt}
\end{table*}

\subsection{Log Evaluator}\label{apex:evaluator_details}

% The GRPO training paradigm enables the model to learn the intrinsic utility logic of historical logs rather than relying on rigid classification patterns. 
% The optimization objective is defined as:

% {
% \small
% \begin{equation}
% \mathcal{J}_{\text{GRPO}}(\theta) = \mathbb{E}_{\genfrac{}{}{0pt}{2}{(q,a) \sim \mathcal{D}}{o_i \sim \pi_{{\text{old}}}}}
% \left[
% \frac{1}{\sum_{i=1}^G |o_i|} \sum_{i=1}^G \sum_{t=1}^{|o_i|} L_{i,t}
% \right]
% \end{equation}
% }

% where $o_i$ denotes the $i$-th sampled output for a given query $q$; $G$ represents the number of sampled outputs per query, $L_{i,t}$ denotes the loss function:

% {
% \small
% \begin{equation}
% L_{i,t} = \min( r_{i,t} \hat{A}_{i,t},\ \text{clip}\left(r_{i,t},\ 1-\varepsilon,\ 1+\varepsilon\right) \hat{A}_{i,t} ),
% \end{equation}
% }

% where $r_{i,t}$ is the importance weight and $\hat{A}_{i,t}$ denotes the normalized advantage.
% In the following, we describe the training data construction process for the Log Evaluator.

\noindent  \textbf{Data Selection and Labeling} \  To construct a high-fidelity training set for the Log Evaluator, we utilized the filtering results from preceding stages (Stage A and Stage B) to identify logs with varying levels of relevance. We designated the top three historical records as positive samples to represent high-utility evidence. To provide the model with discriminative signals, we randomly selected records ranked significantly lower in the initial retrieval as negative instances.
Following the selection, we employed GPT-5.1 to generate the final labels and the underlying reasoning for each sample based on the prompts specified in Table \ref{table:similarity-judge-prompt}. This balanced construction ensures that the model learns to prioritize representative logs that are most conducive to solving the target user query through reinforcement learning.

\begin{table*}[ht]
\centering
\small
\renewcommand{\arraystretch}{1.5}
\begin{tabularx}{\textwidth}{lX}
\toprule
\textbf{Role} & \textbf{System Prompt:} \\
 & You are a Query Similarity Judge. Your task is to determine which historical queries can represent the user's query in terms of capability requirements. \\
\midrule
 & \textbf{Judgment Criteria:} \\
 & 1. \textbf{Skills (S):} Historical query skills should cover user query skills (semantic similarity allowed). \\
 & 2. \textbf{Knowledge (K):} Historical query domains should cover user query domains. \\
\textbf{Template} & 3. \textbf{Difficulty (D):} Historical difficulty must be $\ge$ User difficulty ($D0 < D1 < D2 < D3$). \\
 % \midrule
 & \textbf{User Query \& Historical Pool:} \\
 & \textbf{User Query:} \$\{user\_query\} \\
 & \textit{Decomposition:} S: \$\{user\_skills\}, K: \$\{user\_knowledge\}, D: \$\{user\_difficulty\} \\
 & \textbf{Historical Queries:} \\
  & \$\{query\_a\} \\
  & \$\{query\_b\} \\
  & \$\{query\_c\} \\
 % \midrule
 & \textbf{Instructions:} \\
 & 1. Analyze User Query requirements. \\
 & 2. Compare each historical query (A, B, C) against the User Query based on S, K, and D. \\
 & 3. Determine which queries are valid representatives. \\
 % \midrule
 & \textbf{Output Format (STRICT JSON):} \\
 & \{ \\
 & \quad "thinking": "Brief justification (max 150 words).", \\
 & \quad "valid\_representatives": ["A", "B"] \\
 & \} \\
 \bottomrule
\end{tabularx}
\caption{The prompt template for the Query Similarity Judge. The Log Evaluator also utilizes this exact prompt template for both its training and inference phases}
\label{table:similarity-judge-prompt}
\end{table*}

\noindent  \textbf{Expert Verification Process} \
The integrity of the training data was maintained through a rigorous verification process where three PhD students specializing in Computer Science independently audited each sample. Within this protocol, the experts utilized the generated reasoning trajectories as references and incorporated their own professional judgment to assess whether the resulting log sets provided a valid and representative response to the input query. Adhering to a unanimous consensus requirement, a sample was only incorporated into the final dataset if it received an independent pass score from all three experts, while any samples deemed incorrect were discarded. This collaborative audit effectively ensured high data fidelity to provide a reliable foundation for the subsequent GRPO optimization process.

\section{Experiments Setup}

\subsection{CodaSet Dataset} \label{apex:dataset}

%begin table

% \begin{table*}[ht]
% \centering
% \begin{tabular}{@{}l l l c c l}
% \toprule
% \textbf{Category} & \textbf{Dataset} & \textbf{Task Domain} & \textbf{Training Size} & \textbf{Test Size} \\ \midrule
% ID & MMLU-Pro & Multi-discipline & 6,956 & 2,734 \\
% ID & GSM8K & Math Reasoning & 3,400 & 2,000  \\
% ID & IFEval & Instruction Following & 115 & 415  \\
% ID & BBH & Logical Reasoning & 4,530 & 1,734 \\ \midrule
% % OOD & TruthfulQA & Hallucination/Truth & 0 & 816  \\
% OOD & Math500 & Advanced Math & 0 & 500  \\
% OOD & MT-bench & Multi-turn Chat & 0 & 80  \\
% OOD & MBPP & Python Coding & 0 & 378  \\ \bottomrule
% \end{tabular}
% \caption{\label{apex:dataset_stats}
% Statistical distribution and usage of datasets in CodaSet.}
% \end{table*}

\begin{table*}[ht]
\centering
\begin{tabular}{@{}l l l c c c}
\toprule
\textbf{Category} & \textbf{Dataset} & \textbf{Task Domain} & \textbf{Training Size} & \textbf{Val. Size} & \textbf{Test Size} \\ \midrule
ID & MMLU-Pro & Multi-discipline & 6,756 & 200 & 2,734 \\
ID & GSM8K & Math Reasoning & 3,200 & 200 & 2,000 \\
ID & IFEval & Instruction Following & 130 & 100 & 300 \\
ID & BBH & Logical Reasoning & 4,330 & 200 & 1,734 \\ \midrule
OOD & Math500 & Advanced Math & 0 & 0 & 500 \\
OOD & MT-bench & Multi-turn Chat & 0 & 0 & 80 \\
OOD & MBPP & Python Coding & 0 & 0 & 378 \\ \bottomrule
\end{tabular}
\caption{\label{apex:dataset_stats}
Statistical distribution and usage of datasets in CodaSet. For ID datasets, validation sets are partitioned from the original training sets.}
\end{table*}

%end table

\subsubsection{Dataset Statistics}
The datasets within CodaSet cover a wide spectrum of capabilities, including mathematical reasoning, instruction following, logical deduction, and code generation. For each ID dataset, the data is partitioned into separate training and testing sets. Crucially, to demonstrate that our proposed framework can be seamlessly extended to new domains without the need for additional model training, which effectively allows it to act as a plug-and-play system, we incorporate OOD datasets directly into the evaluation pipeline.
For these OOD tasks, the data is used to evaluate the performance of all base models as well as our proposed model. The detailed statistical breakdown of CodaSet is presented in Table \ref{apex:dataset_stats}.

\subsubsection{Detailed Dataset Descriptions}

\begin{itemize}
    \item \textbf{MMLU-Pro~\cite{mmlu_pro}}: A more challenging extension of the MMLU benchmark, designed with a larger candidate answer space and harder distractors to better assess models’ complex reasoning abilities beyond surface-level knowledge recall.
    
    \item \textbf{GSM8K~\cite{gsm8k}}: A collection of high-quality math word problems that require multi-step reasoning to derive the final answer.
    
    \item \textbf{IFEval~\cite{ifeval}}: A dataset focused on objective verifiable instructions, designed to test the model's ability to strictly adhere to specific formatting constraints and rules.
    
    \item \textbf{BBH (Big-Bench Hard)~\cite{bbh}}: A curated subset of the BIG-bench benchmark composed of particularly challenging tasks, designed to evaluate advanced reasoning capabilities beyond surface-level pattern matching.

    % \item \textbf{TruthfulQA~\cite{truthfulqa}}: A benchmark designed to assess whether models produce truthful responses rather than reproducing common misconceptions, false beliefs, or misleading patterns present in human-generated data.
    
    \item \textbf{Math\_500~\cite{math500}}: A benchmark consisting of challenging mathematical problems across multiple subfields, such as algebra and geometry, used to evaluate advanced mathematical reasoning abilities.
    
    \item \textbf{MT-bench~\cite{mt_bench}}: A multi-turn conversational benchmark that evaluates dialogue performance across diverse categories using an automated large language model–based judge.
    
    \item \textbf{MBPP~\cite{mbpp}}: A benchmark comprising short Python programming problems designed to evaluate programming proficiency, with an emphasis on algorithmic reasoning and code generation.
\end{itemize}

\subsection{Detailed Hyperparameter Configurations}
\label{appendix:hyperparameters}

This section provides the comprehensive hyperparameter settings used for training the Query Deconstructor and the Log Evaluator.
All experiments in this study were conducted on a server cluster equipped with eight NVIDIA H100 GPUs. 

\subsubsection{Query Deconstructor Training (SFT)}
The Query Deconstructor was fine-tuned using a standard Supervised Fine-Tuning (SFT) pipeline. The detailed parameters are listed in Table~\ref{tab:sft_params}.

\begin{table}[ht]
\centering

\begin{tabular}{ll}
\toprule
\textbf{Hyperparameter} & \textbf{Value} \\ \midrule
Backbone Model          & Qwen3-0.6B       \\
Learning Rate           & $2 \times 10^{-5}$ \\
Batch Size       & 128            \\
Optimizer               & AdamW          \\
LR Scheduler            & Cosine         \\
Warmup Ratio            & 0.03           \\
Weight Decay            & 0.1            \\
Max Sequence Length     & 2048           \\
Training Epochs         & 3              \\
Precision               & bf16           \\ \bottomrule
\end{tabular}
\caption{\label{tab:sft_params} Hyperparameters for Query Deconstructor SFT.}
\end{table}

\subsubsection{Log Evaluator Training (RL via verl)}
The Log Evaluator was optimized using the \texttt{verl} framework. The reinforcement learning hyperparameters are summarized in Table~\ref{tab:rl_params}.

\begin{table}[ht]
\centering
\begin{tabular}{ll}
\toprule
\textbf{Hyperparameter} & \textbf{Value} \\ \midrule
Backbone Model          & Qwen3-0.6B       \\
RL Framework            & verl           \\
Actor Learning Rate     & $1 \times 10^{-6}$ \\
Train Batch Size  & 512 \\
Mini Batch Size              & 256             \\
Rollout Samples per Prompt & 8           \\
KL Coefficient          & 0.001           \\
Max Prompt Length       & 4096           \\
Max Response Length     & 2048           \\ \bottomrule
\end{tabular}
\caption{\label{tab:rl_params} Hyperparameters for Log Evaluator RL Training.}
\end{table}

% \subsection{Inference and Decision Parameters}
% Table~\ref{tab:inference_params} lists the thresholds and coefficients used during the multi-stage routing and decision process.

% \begin{table}[ht]
% \centering
% \caption{Inference and Decision Parameters.}
% \label{tab:inference_params}
% \begin{tabular}{lll}
% \toprule
% \textbf{Parameter} & \textbf{Description} & \textbf{Value} \\ \midrule
% $\tau$             & Substage A Score Threshold & 0.5 \\
% $\lambda$          & Empirical Decision Weight   & 0.5 \\
% Embedding Model    & BGE-M3 (Pre-trained)       & Fixed \\ \bottomrule
% \end{tabular}
% \end{table}

\subsection{Detailed Comparison Methods} 
\label{apex:detailed_comparison_methods}
We compare \textbf{DecoR} against the following baselines:

\begin{itemize}
    \item \textbf{Random Router}~\cite{routellm}: Selects a candidate LLM uniformly at random for each incoming query.
    \item \textbf{LLM Router}: A prompt-based routing approach that employs an LLM to select models based on natural-language descriptions of their performance characteristics.
    \item \textbf{RouterDC}~\cite{RouterDC}: A dual contrastive learning-based router that jointly trains query and model embeddings by pulling queries toward suitable models while clustering semantically similar queries in the representation space.
    \item \textbf{EmbedLLM}~\cite{embedllm}: An encoder--decoder framework that learns compact query and model embeddings to predict model--query compatibility, with the router optimized using a binary cross-entropy objective.
    % \item \textbf{SVM Router}:A semantic classifier that dispatches user queries to the most appropriate model or tool by identifying their intent category within a vector space.
    \item \textbf{MODEL-SAT}~\cite{SAT}: A routing method that converts candidate model performance into textual descriptions, which are embedded and processed by a trainable LLM to dynamically select the most suitable model for each query.
    \item \textbf{KNN Router}~\cite{routerbench}: A routing framework that estimates model performance by averaging over the $k$ nearest training examples and routes each query to the LLM with the highest estimated performance.
\end{itemize}

\subsection{Base Models} \label{apx:base_models}

Detailed profiles of the models in our LLM pool are presented below. For a fair and consistent evaluation, inference cost metrics are derived from DeepInfra's pricing effective as of November 17, 2025, a timeframe that coincides with our data collection and experimental phase. 

\begin{itemize}

    \item \textbf{Kimi-K2-Instruct-0905}\footnote{\url{https://deepinfra.com/moonshotai/Kimi-K2-Instruct-0905}} is a Mixture-of-Experts (MoE) model developed by Moonshot AI, possessing a total of 1 trillion parameters with 32 billion active parameters per forward pass. It is optimized for complex instruction following and large-scale language understanding.

    \item \textbf{DeepSeek-V3.1-Terminus}\footnote{\url{https://deepinfra.com/deepseek-ai/DeepSeek-V3.1-Terminus}} is a large-scale hybrid reasoning model featuring 671 billion total parameters and 37 billion active parameters. It supports both thinking and non-thinking modes to balance deep reasoning with inference efficiency.
    
    \item \textbf{DeepSeek-V3.2-Exp}\footnote{\url{https://deepinfra.com/deepseek-ai/DeepSeek-V3.2-Exp}} is an experimental iteration toward next-generation architectures featuring 685 billion parameters. It introduces DeepSeek Sparse Attention to validate optimizations for training and inference efficiency in ultra-long context scenarios.
    
    \item \textbf{Qwen3-235B-A22B-Instruct-2507}\footnote{\url{https://deepinfra.com/Qwen/Qwen3-235B-A22B-Instruct-2507}} is an updated version of the Qwen3 series with 235 billion total and 22 billion active parameters. This version significantly enhances general capabilities in instruction following, logical reasoning, mathematics, and tool usage.
    
    \item \textbf{gpt-oss-120b}\footnote{\url{https://deepinfra.com/openai/gpt-oss-120b}} is an open-weight Mixture-of-Experts (MoE) model from OpenAI with 117 billion parameters. It is designed for high-reasoning tasks, agentic workflows, and general-purpose production use cases.
    
    \item \textbf{gemma-3-27b-it}\footnote{\url{https://deepinfra.com/google/gemma-3-27b-it}} is a 27-billion-parameter instruction-tuned model that supports context windows up to 128k tokens. It features improved reasoning and multilingual capabilities across over 140 languages including support for structured outputs.
    
    \item \textbf{Mistral-Small-3.2-24B-Instruct}\footnote{\url{https://deepinfra.com/mistralai/Mistral-Small-3.2-24B-Instruct-2506}} is a 24-billion-parameter upgrade over the 3.1 release. It demonstrates markedly better instruction following and a more robust function-calling interface while maintaining high performance across text and vision benchmarks.
    
    \item \textbf{gemma-3-12b-it}\footnote{\url{https://deepinfra.com/google/gemma-3-12b-it}} is the 12-billion-parameter variant of the Gemma-3 family. It provides a balanced solution between computational efficiency and advanced reasoning performance for chat-based applications.
\end{itemize}

\section{Experimental Analysis}

% \subsection{Impact of Shifting Threshold \texorpdfstring{$\tau$}{tau} in Substage A} \label{apex:thresh_tau}

% We explore the sensitivity of Substage A to the threshold $\tau$ within the range $[0.1, 0.9]$. As illustrated in Figure \ref{pic:tau}, increasing $\tau$ leads to a gradual improvement in model performance, alongside a corresponding rise in computational cost. In the low $\tau$ regime, performance is relatively limited, though costs remain minimal. As $\tau$ enters the mid-range, performance gains become significant while cost growth remains moderate. However, in the high $\tau$ range, performance tends to saturate while cost increases become more pronounced. Balancing performance enhancement and cost control, we select $\tau = 0.5$ as it achieves an optimal trade-off and serves as our default configuration.

% \begin{figure}[!t]
% \centering
% \includegraphics[width=0.99\linewidth]{latex/pic/tau.png}
% \caption{Trade-off between performance and cost across varying $\tau$ values. The figure illustrates the trends for model accuracy (solid blue line) and normalized cost (dashed red line) as $\tau$ ranges from 0.1 to 0.9. As $\tau$ increases, model performance exhibits an upward trend, accompanied by a corresponding increase in computational cost.}
% \vspace{-4pt}
% \label{pic:tau}
% \end{figure}

\subsection{Case Study} \label{apex:case_study}

To provide readers with a clearer understanding of the internal decision-making logic of DecoR, we present two representative cases in Table \ref{tab:case_study_empty} and Table \ref{tab:case_study_valid} for a comparative analysis:

Case 1: Triggering the Fallback Mechanism (Table \ref{tab:case_study_empty}). This case illustrates the system's rigor when handling domain-specific requirements. The user query involves writing a humorous post about an "Argentinian restaurant." Although the retrieved historical logs overlap partially with the query in terms of formatting requirements, the Log Evaluator accurately identifies the absence of specialized knowledge regarding "Argentinian food culture" and the specific "style imitation" skills required for the target audience. To avoid potential routing errors caused by insufficient prior experience, the system returns an empty set and proactively triggers the fallback mechanism, thereby ensuring performance stability in unfamiliar scenarios.

Case 2: Successful Representative Identification (Table \ref{tab:case_study_valid}). This case demonstrates the system's ability to extract logical commonalities across different contexts. While the user query and the retrieved historical logs involve distinct specific scenarios (such as counting fish versus distributing chocolates), DecoR keenly captures their high consistency in the "arithmetic reasoning" skill dimension and the "D1 difficulty level." This fine-grained, multi-dimensional matching allows the system to effectively reuse historical performance data, enabling an optimal routing decision without the need to blindly invoke high-cost models.

Together, these cases demonstrate how DecoR prevents misjudgments through precise dimensional decomposition and achieves efficient experience reuse when logical cores align, effectively balancing system robustness with cost-efficiency.

\begin{table*}[htbp]
\centering
\small
\renewcommand{\arraystretch}{1.4}
\begin{tabularx}{\textwidth}{l|X}
\toprule
\rowcolor[HTML]{F2F2F2} 
\textbf{Module} & \textbf{Content (Case Study 1: Representative Detection - Empty Output)} \\ \midrule
\textbf{User Query} & Write a funny post for teenagers about a restaurant called "Buena Onda" which serves Argentinian food. Highlight at least three sections of your response in markdown such as *highlighted section*. Mention "Argentinian" in the post. \\ \hline
\textbf{Query Decon.} & \textbf{S:} [writing, creative writing, style imitation, information extraction] \newline \textbf{K:} [general knowledge, food culture, Argentinian cuisine] \quad \textbf{D:} D1 \\ \midrule
\textbf{Historical Logs} & 
\textbf{Log A (ID: 14928):} Generate a forum thread about several people waiting to hear the latest local news... \newline 
\textit{Decon --} \textbf{S:} [information extraction, writing, style imitation]; \textbf{K:} [general knowledge, online communication]; \textbf{D:} D2 \vspace{0.5em} \newline
\textbf{Log B (ID: 14898):} Which of the following is not a fish: salmon or avocado? Answer this easy question first... \newline 
\textit{Decon --} \textbf{S:} [style imitation, writing, information extraction]; \textbf{K:} [general knowledge, nutrition, biology]; \textbf{D:} D2 \vspace{0.5em} \newline
\textbf{Log C (ID: 14843):} Write a dialogue between two people, one is dressed up in a ball gown and the other is dressed down in sweats... \newline 
\textit{Decon --} \textbf{S:} [creative writing, writing, format imitation, dialogue composition]; \textbf{K:} [general knowledge, social interaction]; \textbf{D:} D1 \\ \midrule
\textbf{Log Evaluator} & \textbf{Judgment Reasoning:} Query A involves writing and style imitation with a higher difficulty... but lacks creative writing and humor focus. Query B covers style imitation, writing, and information extraction... but its knowledge domain (nutrition, biology) differs significantly. Query C includes creative writing, writing, format imitation... but lacks information extraction and explicit knowledge of food culture or Argentinian cuisine. Since none fully cover the combined skills and knowledge needed for creative, humorous writing about Argentinian food with format elements, none are adequate representatives. \\ \hline
\textbf{Final Decision} & \texttt{valid\_representatives: []} \newline \textbf{Result:} No representative found. The system triggers the \textbf{fallback mechanism} to ensure performance stability. \\ \bottomrule
\end{tabularx}
\caption{Case Study 1: An instance where the DecoR system fails to find a valid representative in the historical logs. Due to the lack of specialized knowledge (Argentinian cuisine) and specific skill combinations in the retrieved candidates, the Log Evaluator returns an empty set, thereby triggering the fallback mechanism to a high-performance base model.}
\label{tab:case_study_empty}
\end{table*}

\begin{table*}[htbp]
\centering
\small
\renewcommand{\arraystretch}{1.4}
\begin{tabularx}{\textwidth}{l|X}
\toprule
\rowcolor[HTML]{F2F2F2} 
\textbf{Module} & \textbf{Content (Case Study 2: Representative Detection - Valid Output)} \\ \midrule
\textbf{User Query} & There's a Bobbit worm hiding in the bottom of James' aquarium. Every day it eats 2 fish. After two weeks, James adds 8 more fish to the aquarium. A week later, he discovers the Bobbit worm. If the aquarium had 60 fish to start with, how many does it have when James discovers the Bobbit worm? \\ \hline
\textbf{Query Decon.} & \textbf{S:} [mathematics, logical inference, arithmetic reasoning] \newline \textbf{S\_reason:} The query requires calculating the number of fish over time using arithmetic and logical reasoning. \newline \textbf{K:} [basic mathematics, problem solving] \quad \textbf{D:} D1 \\ \midrule
\textbf{Historical  Logs} & 
\textbf{Log A (ID: 2333):} Susan has 3 fish tanks to fill. 1 fish tank contains 7 goldfish and 8 beta fish. The second fish tank contains twice as many fish as the first tank... \newline 
\textit{Decon --} \textbf{S:} [mathematics, logical inference, arithmetic]; \textbf{K:} [basic mathematics]; \textbf{D:} D1 \vspace{0.5em} \newline
\textbf{Log B (ID: 5454):} Hank gave his wife, Delphine, a box of 24 chocolates... On the first day, Delphine ate 4 chocolates. On the second day, she ate 3 less than twice as many... \newline 
\textit{Decon --} \textbf{S:} [mathematics, arithmetic reasoning, logical inference]; \textbf{K:} [basic arithmetic, problem solving]; \textbf{D:} D1 \vspace{0.5em} \newline
\textbf{Log C (ID: 5321):} Of the 3 friends, Harry has 4 times as many fish as Joe, and Joe has 8 times as many fish as Sam does. If Sam has 7 fish... \newline 
\textit{Decon --} \textbf{S:} [basic arithmetic, mathematics, logical inference]; \textbf{K:} [basic mathematics]; \textbf{D:} D1 \\ \midrule
\textbf{Log Evaluator} & \textbf{Judgment Reasoning:} All three historical queries involve arithmetic reasoning, logical inference, and basic mathematics, covering the skills and knowledge required by the user's query. Their difficulty levels are the same as the user's (D1), satisfying the requirement for difficulty. Each historical query involves multi-step calculations with similar or overlapping reasoning and problem-solving concepts, making them suitable representatives of the user's query. \\ \hline
\textbf{Final Decision} & \texttt{valid\_representatives: ["A", "B", "C"]} \newline \textbf{Result:} Successfully matched representatives. The system can leverage the prior performance data of these logs to make an informed routing decision. \\ \bottomrule
\end{tabularx}
\caption{Case Study 2: An instance where the DecoR system identifies valid representatives. Unlike Case Study 1, the retrieved logs here share identical Skill sets (arithmetic reasoning, logical inference) and Difficulty levels (D1) with the user query. The Log Evaluator confirms that the reasoning patterns are sufficiently similar, allowing the system to reuse historical performance scores instead of triggering the fallback mechanism.}
\label{tab:case_study_valid}
\end{table*}

\end{document}